\documentclass[pdflatex,sn-mathphys-num]{sn-jnl}

\usepackage{graphicx}%
\usepackage{multirow}%
\usepackage{amsmath,amssymb,amsfonts}%
\usepackage{amsthm}%
\usepackage{mathrsfs}%
\usepackage[title]{appendix}%
\usepackage{xcolor}%
\usepackage{textcomp}%
\usepackage{manyfoot}%
\usepackage{booktabs}%
\usepackage{algorithm}%
\usepackage{algorithmicx}%
\usepackage{algpseudocode}%
\usepackage{listings}%

\usepackage{makecell}
\usepackage{subcaption}

\theoremstyle{thmstyleone}%
\theoremstyle{thmstyletwo}%

\theoremstyle{thmstylethree}%

\begin{document}

\title[Article Title]{Physical simulation of Marsupial UAV-UGV Systems Connected by a Variable-Length Hanging Tether}


\author*[1]{\fnm{José E.} \sur{Maese}}\email{jemaealv@upo.es}

\author[1]{\fnm{Fernando} \sur{Caballero}}\email{fcaballero@upo.es}

\author[1]{\fnm{Luis} \sur{Merino}}\email{lmercab@upo.es}

\affil[1]{\orgdiv{Service Robotics Lab}, \orgname{Universidad Pablo de Olavide}, \orgaddress{\city{Seville}, \postcode{41013}, \country{Spain}}}


\abstract{This paper presents a simulation framework able of modeling the dynamics of a hanging tether with adjustable length, connecting a UAV to a UGV. The model incorporates the interaction between the UAV, UGV, and a winch, allowing for dynamic tether adjustments based on the relative motion of the robots. The accuracy and reliability of the simulator are assessed through extensive experiments, including comparisons with real-world experiment, to evaluate its ability to reproduce the complex tether dynamics observed in physical deployments. The results demonstrate that the simulation closely aligns with real-world behavior, particularly in constrained environments where tether effects are significant. This work provides a validated tool for studying tethered robotic systems, offering valuable insights into their motion dynamics and control strategies.}




\maketitle

\renewcommand{\thefootnote}{\relax}
\footnotetext{This work was partially supported by the grants: 1) INSERTION PID2021-127648OB-C31, and 2) RATEC PDC2022-133643-C21, funded by MCIN/AEI/ 10.13039/501100011033 and the “European Union NextGenerationEU/PRTR”}
\renewcommand{\thefootnote}{\arabic{footnote}}


\section{Introduction} 

Marsupial systems, where unmanned aerial vehicles (UAVs) are deployed from unmanned ground vehicles (UGVs) to combine their complementary capabilities, are emerging as innovative solutions for a broad spectrum of complex applications. These systems offer the unique capability of operating on challenging and expansive terrain. In the case of tethered configurations, the UAV's operational time can be significantly extended through continuous power supply via a tether connected to the UGV, effectively overcoming the limited flight duration that typically restricts small and medium-sized UAVs to a few tens of minutes. This enhancement considerably increases the mission range and operational flexibility. Prominent applications of this technology include search and rescue operations, critical infrastructure inspections, and advanced military missions. The UAV's ability to monitor and coordinate from the air while the UGV performs specific tasks on the ground enables effective integration of aerial and terrestrial capabilities, thus optimizing the UAV's operational duration.

However, tethered configurations inherently limits maneuverability. To overcome this constraint, advanced planning algorithms and appropiate simulation framework are required. Developing these algorithms to coordinate these marsupial systems presents significant challenges. Designing controls that precisely synchronize the actions of UAVs and UGVs requires facing challenges related to communication, synchronization, and data fusion. The complexity increases when considering the tethers connecting these vehicles, especially in modeling and controlling the dynamics of hanging or partially taut tethers, a key aspect not yet considered by existing simulators in the state-of-the-art.

\begin{figure}[]
    \centering  
    \includegraphics[width=0.8\columnwidth]{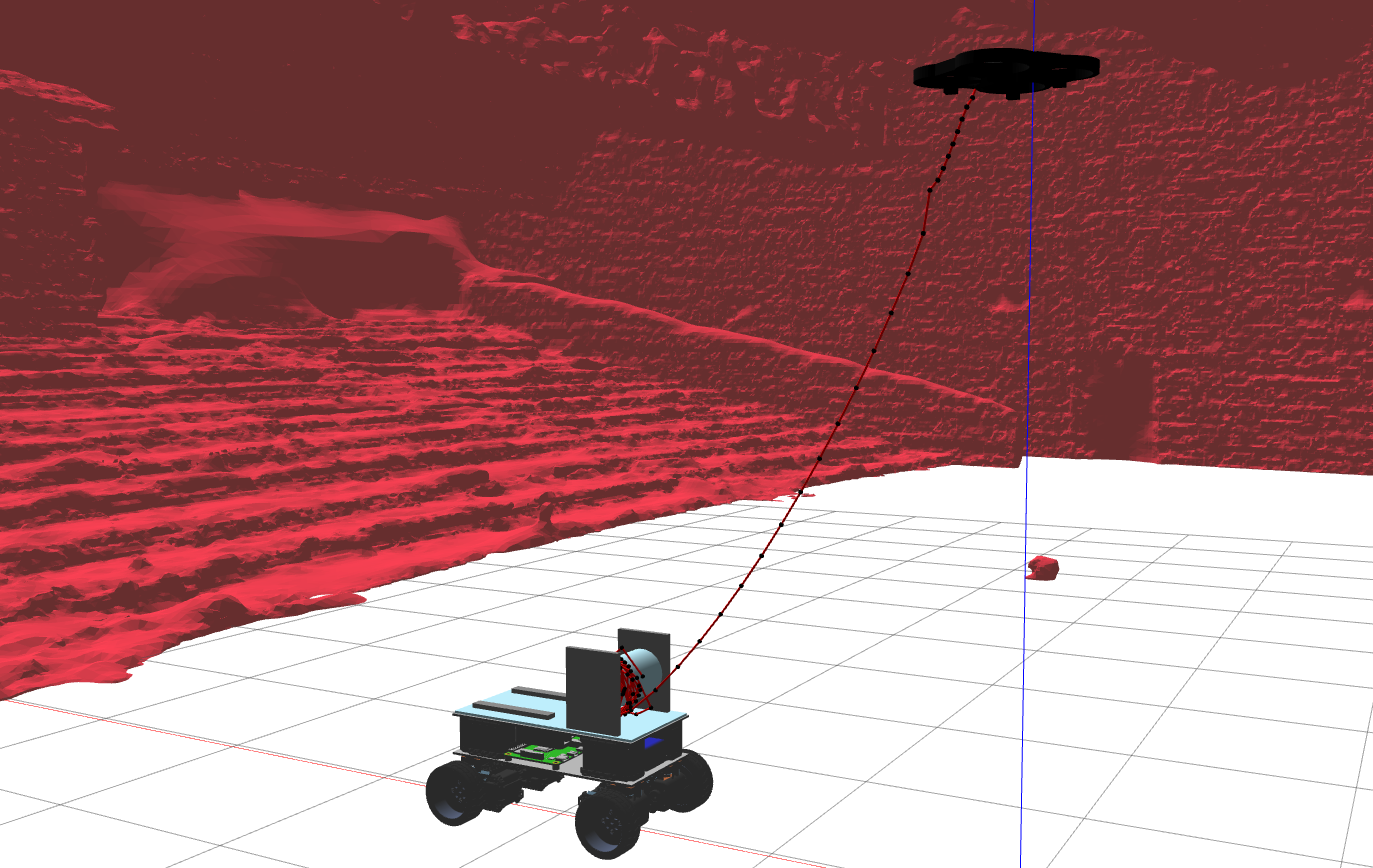}
    \caption{Example of the proposed marsupial system simulator deployed in a theatre model using Gazebo.}
    \label{figure:initial_image}
\end{figure}

The authors of this paper have previously tackled the problem of trajectory planning for a marsupial system where a UAV and UGV are connected by an adjustable hanging tether, as presented in \cite{Martinez-Rozas2023}. The paper introduces a solution based on the Rapidly-exploring Random Trees (RRT) and nonlinear optimization to compute collision-free paths for both the UAV and UGV while considering the tether's dynamics and environmental obstacles. While this approach was effective for planning, it lacked a detailed physics-based simulation of tether behaviour under various real-world conditions. Consequently, such research highlighted the need for accurate simulation of tethered UAV-UGV, and how crucial it would be for overcoming the practical difficulties associated with real-world testing. 

This paper proposes and studies the viability of a physics simulator able of modelling the dynamics of a hanging tether of adjustable length connecting a UAV to an UGV, enabling researchers and developers to test and optimize their algorithms in a controlled environment prior to physical implementation. This not only would minimize the costs and risks associated with physical testing, but also would accelerate the development process. This simulation is validated using the work  on \cite{Martinez-Rozas2023}, by leveraging the trajectories generated by the algorithm presented there to check the usefulness and robustness of our ROS 2-based simulator framework. Specifically, we replicated the same trajectories from their real-world experiments (Fig. \ref{figure:initial_image}), demonstrating how our simulator handles hanging-tether dynamics in real scenarios.
 
The paper is structured as follows: Section \ref{sec:related-work} reviews the existing literature on marsupial systems and tether modelling. Section \ref{sec:gazebo-sim} describes the simulator architecture, the models used, and the characteristics of the simulation environment in Gazebo. The metrics of the simulator are presented in Section \ref{sec:sim-eval}, there, the metrics and the data obtained are detailed. Section \ref{sec:val-exp} evaluates the simulator’s capabilities through experiments on tether model accuracy, system behavior, and computational performance, including a comparison with real-world data. Finally, Section \ref{sec:conclusions} presents the conclusions and future work, summarizing the findings, and discusses potential directions for future research.


\section{Related Work}
\label{sec:related-work}

 \begin{figure}[]
      \centering
      \includegraphics[width=1.0\columnwidth]{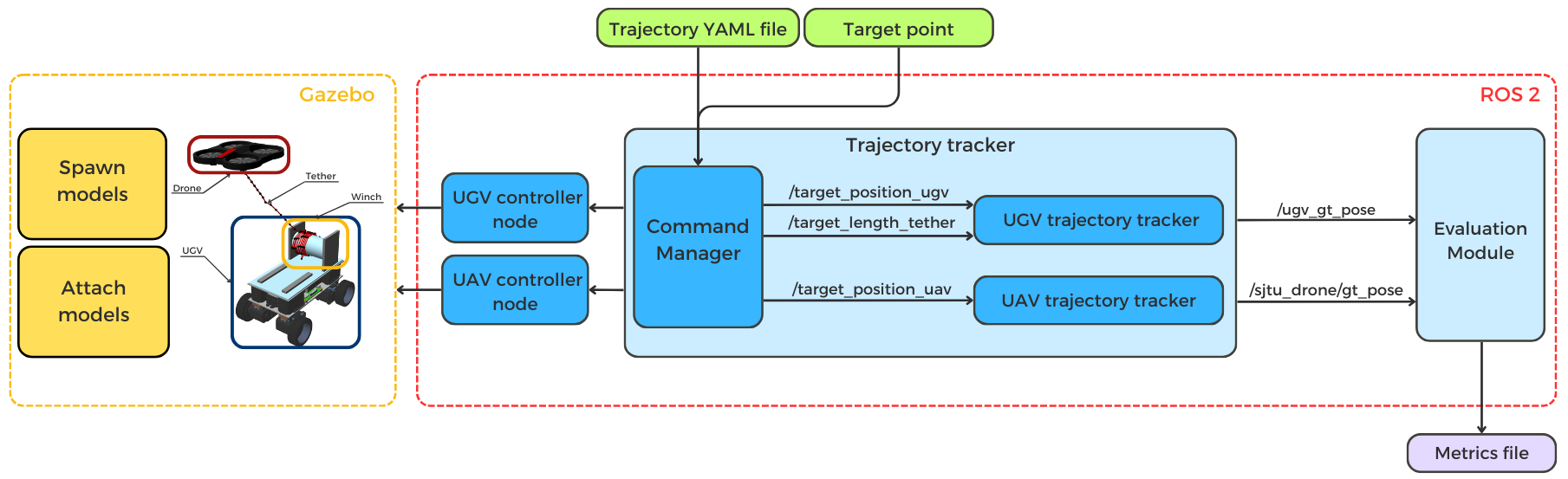}
      \caption{Diagram of the Marsupial Simulator. The core modules of simulator are in blue. The base robotics simulator modules are depicted in light yellow color. Green boxes indicate the input data information. Output evaluation metrics are indicated in light purple color.}
      \label{figure:simulator_estructure}
   \end{figure}


Marsupial robotic systems, where a UAV is deployed from a UGV, have been the subject of various studies in recent years. These systems offer unique advantages for applications in complex environments, where the ground and aerial capabilities can be leveraged for increased versatility. Early works, such as \cite{Papachristos2014}, assumed a taut tether, which restricted the UAV’s maneuverability in obstacle-rich environments. In contrast, \cite{Martinez-Rozas2023} presented a path and trajectory planning solution specifically for UAV-UGV systems connected by an adjustable tether. Their approach allows the tether to remain slack, providing the UAV with greater flexibility and overcoming some of the limitations found in prior studies.

In real-world applications, marsupial UAV-UGV systems have proven invaluable in hazardous and confined environments. For instance, the CSIRO team used such configurations during the DARPA Subterranean Challenge to perform autonomous exploration and mapping \cite{Hudson2022}. This mission illustrated the practical benefits of using autonomous navigation for multi-agent collaboration. Another field study, \cite{Oh2006}, explored the landing of autonomous helicopters using tethers as guiding mechanisms. This exemplifies the relevance of tethered systems in enhancing UAV-UGV cooperation, particularly in safety-critical operations.

However, the development and validation of a tethered system is a complex and risky task, involving two robots with a physical link that might produce significant perturbations in the UAV control subsystem. Realistic simulators stand as a great tool to speed up such development, but simulating tethered systems, especially when considering hanging-tether dynamics, also presents several challenges. Early works on simulators, such as \cite{Bulic2022}, focused on accurately representing the mechanical properties of tethers using finite element models. This approach, while useful for studying the structural behavior of tethers under different loads, lacks the integration with the kinematic control of the UAV-UGV systems, which is essential for real-world applications. In contrast, our approach prioritizes system-level simulation, allowing for the analysis of robot interactions, control strategies, and tether behavior in a comprehensive and operationally relevant manner.

In the context of robotic simulators, Gazebo has become a popular platform for testing UAV-UGV marsupial systems. The work described in \cite{Martinez-Rozas2023} does not simulate the physical dynamics of the tether in Gazebo. Instead, their approach models the tether using a mathematical catenary representation to compute the configuration and adjust the tether’s length during the robots' movement. While this provides an effective method for trajectory optimization, the simulation does not include the tether’s physical behavior, such as slack, in real-time. Similarly, packages such as \texttt{sitl\_gazebo} \cite{Huang2023}, which extend Gazebo’s capabilities for UAV simulations, offer practical implementations for tethered robotic systems, although these are often limited to specific configurations and lack the flexibility required for comprehensive tether management.

Another notable contribution is \cite{Caruso2021}, which explores the kinematic control of tethered robots, treating the tether as an additional degree of freedom. This approach improves control over the length of the tether, but does not fully integrate the complex dynamics of the tether in environments where it can sag or interact with obstacles. By contrast, recent works, such as \cite{Laranjeira2020}, model tether dynamics in underwater systems using a catenary-based visual servoing approach, which could potentially be adapted for aerial-ground systems in confined spaces.

Simulations of tethered systems are crucial for evaluating algorithms before deployment in real-world settings. The ability to test various control strategies, such as the tether-aware path planning method in \cite{Petit2022}, allows researchers to fine-tune their solutions without the risks associated with physical experimentation. In our case, the presented ROS 2-based simulator framework offers a unique capability by focusing on the real-time dynamics of the entire UAV-UGV system. It incorporates both manual and automated trajectory control, with dynamic adjustments of the tether's length and slack, enabling a flexible and comprehensive testing environment. Overall, advances in Gazebo-based simulators, combined with more complex tether dynamics models, are paving the way for more accurate and efficient marsupial robotic systems.


\section{Marsupial System Simulator using Gazebo} \label{sec:gazebo-sim}

The architecture of the marsupial UAV-UGV simulator system consists of multiple core components that interact to simulate the tethered robots and their behavior. A detailed representation of this architecture is provided in Fig. \ref{figure:simulator_estructure}. The system is built using ROS 2 and Gazebo, where each element serves a specific role in the simulation. 

To understand how the core components work together during the simulator's operation, it is crucial to examine the key steps involved. The system can operate under both manual and autonomous control, with the main steps outlined below:

\begin{itemize}
\item \textbf{Model Initialization}: Gazebo acts as the primary simulation environment, where the physical models are instantiated. The simulation begins by spawning the UGV, UAV, and tether in the Gazebo environment. The UAV is placed on a platform on the UGV, while the tether is initialized in a coiled configuration around the winch. The tether is connected at both ends to the UAV and winch, and this setup ensures that the system is ready for simulation. Within this environment, ROS~2-based software modules are responsible for managing the system's operation.

\item \textbf{Trajectory Tracking}: The trajectory tracking module allows easy interaction with the system, acting as the command interface of the tethered system. It offers two different commanding approaches: (1) input files formatted as YAML files that include the waypoints and the reference tether lengths for processing, and (2) specific destination points via ROS messages, enabling dynamic control of the UAV and UGV during the simulation without the need to provide a full trajectory in advance. The trajectory tracker node sends control commands to the UAV and UGV (including its winch) guiding them through their assigned waypoints. As the UAV and UGV move, the winch dynamically adjusts the length of the tether, either winding or unwinding it to maintain the proper slack based on the relative positions of the two robots. The simulation framework is designed to be flexible and modular, allowing users to integrate their own trajectory tracking algorithms or control modules. 

\item \textbf{Controllers}: Each robot, the UAV and the UGV, has its own controller that receives destination points from the trajectory tracking module and executes the required movements. The tether control is embedded in the UGV. They have been tuned to closely resemble the behaviour of the robots used in the real-world experiments (see below), but they can be adapted to follow different dynamics. This modular design allows researchers to replace default components with their own implementations, facilitating the testing and evaluation of various control strategies within the same simulation environment.

\item \textbf{Evaluation and Data Recording}: In addition to the control functions, the system includes an Evaluation Module that collects experimental data throughout the simulation. Throughout the simulation, the Evaluation Module records relevant data from the ROS~2 topics. This data includes the ground truth pose of both the UAV and UGV, as well as the deployed or collected tether length. This information is critical for assessing the system’s performance. The Evaluation Module compiles the data into an output metrics file. Once the simulation is completed, the evaluator processes this data to generate necessary metrics such as trajectory accuracy, tether behavior, and overall system stability. This file serves as a comprehensive record of the simulation’s outcomes, enabling researchers to analyze the behavior and interactions of the tethered UAV-UGV system in detail. The results are saved in a metrics file for further analysis and validation.

\end{itemize}

The simulator is designed to be flexible and modular, allowing for easy modifications to the UAV, UGV, and tether models as needed. The complete documentation, including how to modify the models and run the simulations, as well as the code, is available on the project's GitHub repository\footnote{\url{https://github.com/robotics-upo/marsupial_simulator_ros2}}.

\subsection{Models}

\begin{figure}[]
    \centering
    \includegraphics[width=0.7\columnwidth]{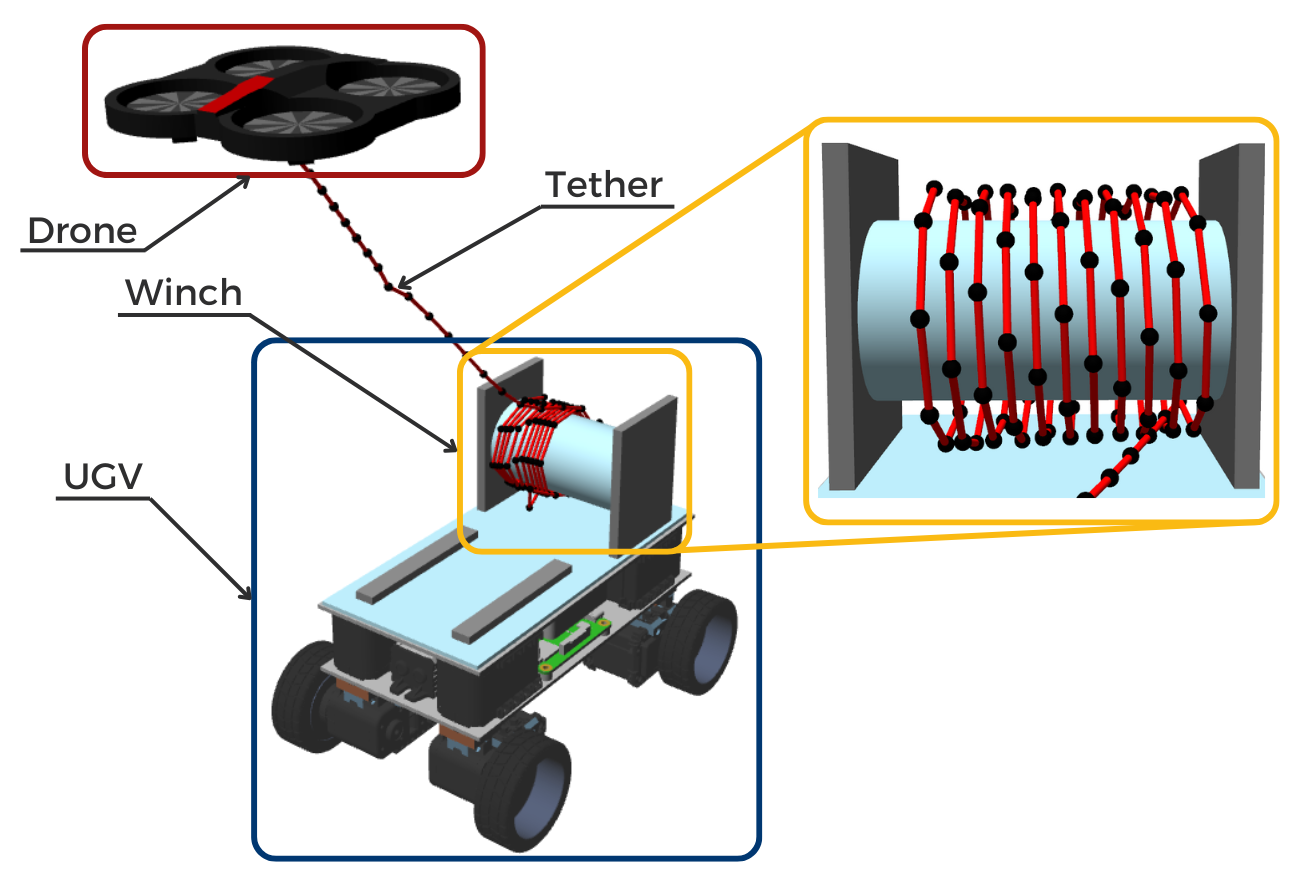}
    \caption{Models used in Gazebo simulation.}
    \label{figure:marsupial_models}
\end{figure}

In the simulation environment, different models are used to accurately represent the components of the marsupial robotic system, such as the UAV, UGV, winch, and tether (Fig. \ref{figure:marsupial_models}). Each of these models have been selected, modified, or developed to ensure proper functionality and realistic behavior within the Gazebo simulator.

\begin{table}[]
\caption{Default tether parameters}\label{tab:tether_parameters}
\begin{tabular}{lcc}
\toprule
\multicolumn{1}{c}{\textbf{Parameters}} & \textbf{Values}  & \textbf{Units} \\
\midrule
\multicolumn{3}{c}{\textit{Common parameters}} \\ 
Radius of each section    & 0.004 & m      \\ 
Radius of the joint       & 0.009 & m      \\ 
Mass of each section      & 0.01  & kg     \\ 
Damping                   & 0.05 & Ns/m    \\ 
Spring stiffness          & 0.01  & N/m    \\ 
\midrule
\multicolumn{3}{c}{\textit{Coiled tether parameters}} \\
Number of elements        & 123  & -       \\ 
Element length            & 0.15  & m      \\ 
Helix radius              & 0.14 & m       \\ 
\midrule
\multicolumn{3}{c}{\textit{Uncoiled tether parameters}} \\
Number of elements        & 10   & -       \\ 
Element length            & 0.05 & m       \\ 
\botrule
\end{tabular}
\end{table}

The UAV model \cite{Luqman2024} is a quadrotor equipped with various sensors, including cameras, IMU, sonar, Lidar, and GPS, which provide essential data for control and navigation. The drone is capable of executing basic flight commands, such as take-off, landing, and directional movement, and can switch between position control and velocity control modes. Position control allows the drone to move to specific poses in 3D space, while velocity control offers more dynamic and flexible movements based on the drone's velocity inputs. In addition, the model supports state feedback, publishing ground-truth data about its position, velocity, and acceleration, which can be used for experiments focused on control and is essential for precise evaluation during simulation. To integrate the drone into the marsupial system, its thrust along the Z-axis has been increased to compensate for the additional weight of the tether. This modification ensures stable flight dynamics even with the tether attached. Other than this modification, the drone's structure and performance characteristics remain unchanged from the original model.

The UGV model used in this simulation is derived from a holonomic platform sourced from open-source software \cite{4WSrobot}. This model allows for omnidirectional movement, providing enhanced manoeuvrability in tight or complex environments. The UGV's holonomic drive system enables it to move laterally, diagonally, and rotate in place, making it ideal for coordinating movements with the UAV while managing the tether in dynamic scenarios. Parameters such as motor power and steering response have been optimized to ensure smooth interactions with the winch and tether throughout the simulation.

The winch, custom-developed and fully integrated with the UGV, operates as a unified system. It includes a rotating cylinder that dynamically winds or unwinds the tether based on the relative positions of the UAV and UGV or the reference tether lengths, maintaining adequate slack during operation. In addition, the winch features a forward-mounted platform designed as a stable surface for the UAV to take off. The control of the winch is managed by the UGV's controller module, ensuring synchronized tether management and smooth UAV-UGV coordination during autonomous or manual control scenarios. This integrated design provides a cohesive simulation of the tether dynamics and marsupial robotic systems, enhancing the realism of aerial and ground interactions.

The tether model used in this simulator is adapted from an open-source repository \cite{Huang2023}, specifically designed for airborne systems in Gazebo. The tether is made up of multiple connected elements, each defined by parameters such as length, radius, mass, and joint stiffness, enabling realistic dynamic behaviours. The default parameters used in the system validation can be found in Table \ref{tab:tether_parameters}. These parameters, including damping and friction, can be adjusted to suit the needs of the simulation. The model is initialized with a fully retracted configuration around the winch and dynamically adjusts its length during operation. This ensures that the tether behaves realistically as it unwinds or winds based on the relative movements of the UAV and UGV. The tether's flexibility and ability to adapt its physical properties, such as stiffness and damping, make it essential for simulating complex tethered systems. Additionally, its attachment points to both the UAV and winch are critical for maintaining consistent slack throughout the UAV-UGV coordination. 

\begin{figure}[]
    \centering
    \includegraphics[width=0.9\columnwidth]{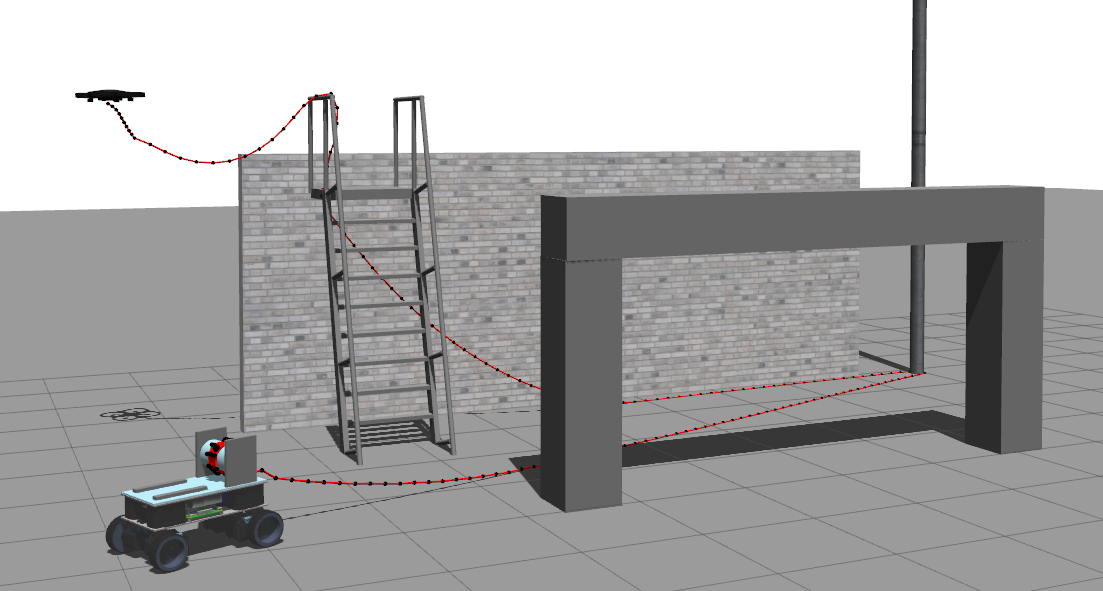}
    \caption{Example of tether-obstacle interaction in Gazebo.}
    \label{figure:tether_obstacles}
\end{figure}

Beyond its dynamic adjustments, the tether model also interacts with the environment through Gazebo's physics engine, which computes contact forces based on material properties and collision geometry. These interactions create local tension variations that propagate along the tether, potentially affecting UAV and UGV dynamics. As illustrated in Fig. \ref{figure:tether_obstacles}, Gazebo naturally handles tether-obstacle interactions, enabling realistic responses to environmental constraints. To assess these effects, the simulator includes six scenarios with different architectural structures, enabling analysis of tether behavior under varied geometric conditions.


\section{Metrics for Simulation Evaluation}
\label{sec:sim-eval}

In order to create a comprehensive and scalable evaluation system, a wide range of simulation data has been collected. The primary objective of this evaluation is to assess the feasibility of the simulator as a tool for comparing trajectory tracking algorithms. A key aspect of this is demonstrating the plausibility of the simulated physical dynamics, particularly those governing the behavior of the tether. 

\subsection{Metrics Description}

This dataset includes information about the velocity and position of both the UGV (along with its associated winch) and the UAV, as well as the length of the tether. The data has been chosen to cover the key aspects of the system's dynamics, focusing on movement and tether behavior. The specific metrics derived from this data are summarized in Table~\ref{tab:metrics}.

The total distance travelled by both the UGV and UAV is approximated by summing the Euclidean distances between each pair of successive points in their respective trajectories. Given a series of positions \( \ p_i = (x_i, y_i, z_i) \) for either the UGV or UAV, the total distance travelled, \( D \), is calculated as:

\begin{equation}
    D = \sum_{i=1}^{n-1} \sqrt{(x_{i+1} - x_i)^2 + (y_{i+1} - y_i)^2 + (z_{i+1} - z_i)^2}
\end{equation}

The length of tether released, \( L_{released} \), is calculated as the sum of the positive increments in the tether length between two consecutive moments in time. If \( L_i \) represents the tether length at time \( t_i \), then:

\begin{equation}
    L_{released} = \sum_{i=1}^{n-1} \max(0, L_{i+1} - L_i)
\end{equation}

Conversely, the length of tether collected, \( L_{collected} \), is the sum of the negative increments in tether length, which corresponds to when the tether is being retracted:

\begin{equation}
    L_{collected} = \sum_{i=1}^{n-1} \max(0, L_i - L_{i+1})
\end{equation}

In addition to the primary metrics, we assessed the fidelity of the simulated tether in replicating a real catenary curve. This evaluation focuses on the impact of tether discretization on simulation accuracy, specifically examining how varying the length of the tether elements affects the simulation's ability to approximate the behavior of an ideal catenary.



Beyond the tether shape itself, these metrics also provide insights into how the UGV and UAV navigate the environment, as well as how the tether behaves as it is released and retracted during motion. Ultimately, the metrics presented here are crucial for verifying that the tether’s behavior remains consistent with expected physical properties, such as rigidity, elasticity, and its interaction with both the UGV and UAV throughout operation.

\begin{table}[]
\centering
\caption{Simulation Metrics and Graphs} \label{tab:metrics}
\begin{tabular}{lcc}
\toprule
\multicolumn{3}{c}{\textbf{Metrics}}                                                       \\ 
\midrule
\multicolumn{1}{c}{\textbf{Metric}} & \textbf{Description} & \textbf{Units}                \\
\midrule
\texttt{sim\_time}          & \multicolumn{1}{l}{Simulation duration}                & s   \\ 
\texttt{targets}            & \multicolumn{1}{l}{Number of target points}            & -   \\ 
\texttt{dist\_ugv}          & \multicolumn{1}{l}{Distance traveled by UGV}           & m   \\ 
\texttt{dist\_uav}          & \multicolumn{1}{l}{Distance traveled by UAV}           & m   \\ 
\texttt{tether\_released}   & \multicolumn{1}{l}{Tether released}                    & m   \\ 
\texttt{tether\_collected}  & \multicolumn{1}{l}{Tether collected}                   & m   \\ 
\texttt{cat\_error}         & \multicolumn{1}{l}{Mean catenary error}                & m   \\ 
\hline
\midrule
\multicolumn{3}{c}{\textbf{Graphs}}                                                        \\ 
\midrule
\multicolumn{1}{c}{\textbf{Graph}} & \textbf{Description} & \textbf{Units}                 \\ 
\midrule
\texttt{pos\_3D}            & \multicolumn{1}{l}{3D position visualization}          & m   \\ 
\texttt{uav\_pos}           & \multicolumn{1}{l}{UAV position on axes}               & m   \\ 
\texttt{ugv\_pos}           & \multicolumn{1}{l}{UGV position on axes}               & m   \\ 
\texttt{uav\_vel}           & \multicolumn{1}{l}{UAV velocity}                       & m/s \\ 
\texttt{ugv\_vel}           & \multicolumn{1}{l}{UGV velocity}                       & m/s \\ 
\texttt{tether\_len}        & \multicolumn{1}{l}{Released length of the tether}      & m   \\ 
\texttt{min\_dist}          & \multicolumn{1}{l}{Minimum distance to obstacles}      & m   \\ 
\botrule
\end{tabular}
\end{table}

\subsection{Computed Metrics Data}

All relevant data was automatically recorded and stored in a set of CSV files for easy access. Each file contains the necessary information for post-simulation analysis. The data is organized in a tabular format, starting with headers that identify the recorded metrics, followed by the corresponding values for each time step.

These output files include general data related to the UAV, UGV, and tether dynamics. Specifically, one file captures the UAV’s position, velocity, and target information, another contains the UGV’s kinematic data and winch details, while the tether data file provides the position information necessary for evaluating its behavior. These files facilitate a comprehensive analysis of the system's performance, allowing for detailed evaluation of the simulation's accuracy and dynamics over time.

This structure enables the visualization and plotting of various metrics, helping to track the behavior of key variables at each time step of the simulation, as well as in specific scenarios or events.


\section{Validation Experiments}
\label{sec:val-exp}

To evaluate the performance of the simulator and validate its capabilities, we designed a series of scenarios that assess various aspects of the marsupial UAV-UGV system, including stability, tether dynamics, winch operation, and coordination between aerial and ground vehicles under different circumstances. The scenarios are divided into four categories: Tether Model Evaluation, aimed at assessing the accuracy of the simulated tether against a real catenary curve; Simulated Scenarios, focused on testing and verifying the system's general functionalities; Real Scenarios, that replicate real-world experiments to validate the simulator in practical applications; and Performance Evaluation, dedicated to computational efficiency and real-time capabilities of the simulator under different conditions.

The documentation in our GitHub repository details how to configure new environments, robot initial poses, motion trajectories, and tether parameters, allowing researchers to adapt the framework to a wide range of experimental setups. Additionally, a summary of the simulation and real-world experiments is presented in the accompanying video\footnote{\url{https://youtu.be/ZLLDROIaHV0}}, providing a visual overview of the system’s capabilities. This flexibility ensures the simulator remains a valuable and continually evolving tool for marsupial UAV-UGV research.

\subsection{Tether Model Evaluation}

To validate the accuracy of the simulated tether against a real catenary, we designed an experiment focusing on the tether's discretization effect. In these experiments, both the UAV and UGV were positioned approximately 5, 10 and 15 meters apart, and the tether length was set to be 20\% longer than the Euclidean distance between them to introduce realistic slack.

We simulate the tether using four different lengths of elements: 0.05 m, 0.10 m, 0.15 m and 0.20 m. For each simulation, we recorded the length of each tether element and calculated the averaged error between the positions of the simulated tether elements and those of the theoretical catenary curve under the same conditions. This error measurement provides insights into how the granularity of the tether model influences the simulation's ability to accurately replicate real-world tether dynamics.

\begin{table}[t!]
\centering
\caption{Error between the simulated tether and the theoretical catenary for tether lengths of 5 m, 10 m, 15 m, and different element lengths (mean values)} \label{tab:catenary_error}
\begin{tabular}{cccccc}
\toprule
\multicolumn{1}{c}{\textit{\makecell{Element \\ length (m)}}} & \textit{$err_5$ (\%)} & \textit{$err_{10}$ (\%)}  & \textit{$err_{15}$ (\%)} & \textit{mean (\%)} \\ 
\midrule
0.05 &  0.942 & 0.268 & 0.269 & 0.493 \\ 
0.10 &  0.332 & 0.214 & 0.232 & 0.259 \\ 
0.15 &  0.675 & 0.323 & 0.292 & 0.430 \\ 
0.20 &  1.600 & 0.490 & 0.552 & 0.881 \\ 
\botrule
\end{tabular}
\end{table}

As shown in Table~\ref{tab:catenary_error}, the results indicate that the performance is realistic across all tested cases, with averaged errors below 1\% of the tether length. We can see how, in general, as the length of the element increases, the simulation error with respect to the theoretical model also increases, as expected. However, this effect does not hold when the element length is 0.05 m. We think this behaviour is related with the increase of the simulation complexity, setting Gazebo's solver close to its limits. This is due to the reduced computational load when simulating fewer (longer) elements. 

It is important to note that the region around the winch (highlighted in yellow in Fig.~\ref{figure:marsupial_models}) exhibits more chaotical tether dynamics due to the high concentration of elements in a small area. This effect is inherent to the discrete nature of the simulation and cannot be directly controlled, leading to small fluctuations in the tether's appearance near the winch. However, this is purely a visual artifact and does not impact the overall behaviour of the catenary, which remains accurately modeled and is the primary focus of our simulation. Additionally, the winch model used in our simulation does not support tether element lengths greater than 0.2 m due to mechanical constraints. Despite varying the element lengths up to this limit, the precision of the simulation was not affected significantly. This suggests that even with larger element lengths within the supported range, the simulation can maintain a high degree of accuracy in approximating the real catenary curve.

\begin{figure}[]
    \centering
    \begin{subfigure}{0.48\columnwidth}
        \centering
        \includegraphics[width=\linewidth]{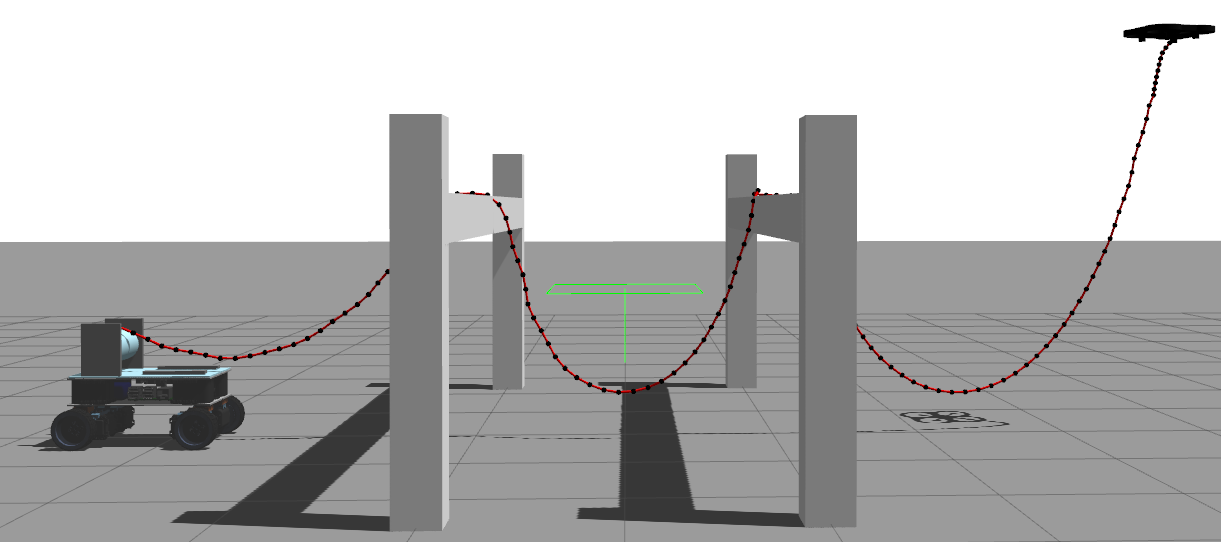}
        \caption{Collision with a hanging tether.}
        \label{fig:collision_a}
    \end{subfigure}
    \hfill
    \begin{subfigure}{0.48\columnwidth}
        \centering
        \includegraphics[width=\linewidth]{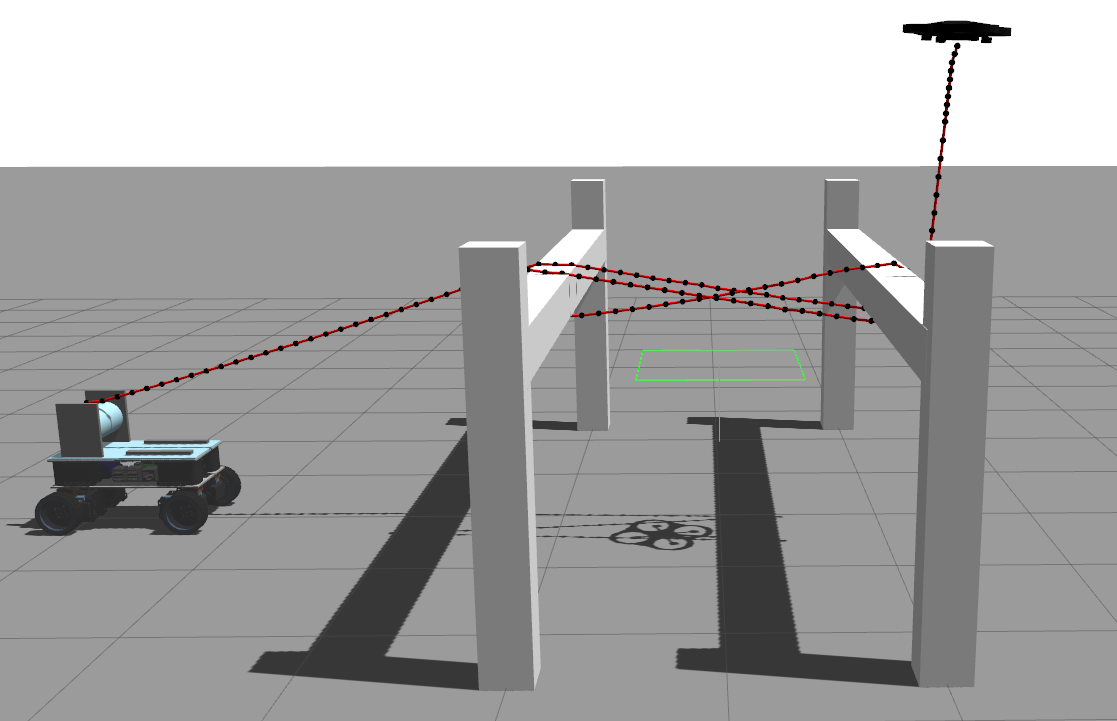}
        \caption{Collision with a taut tether.}
        \label{fig:collision_b}
    \end{subfigure}
    \caption{Examples of tether-obstacle collisions with varying degrees of tension.}
    \label{figure:collisions}
\end{figure}


Beyond comparing the simulated tether to a theoretical catenary, we also conducted experiments to examine how it interacts with the environment. These tests were designed to confirm the simulator’s ability to capture contact dynamics, including collisions with obstacles, wrapping around structures, and adjusting the tether's tension in response. Thus,  Fig.~\ref{fig:collision_a} illustrates a scenario where the tether remains slack while suspended between two obstacles. In contrast, Fig.~\ref{fig:collision_b} shows the tether becoming entangled around both obstacles due to the UAV’s trajectory, significantly increasing tension. These interactions influence the overall system behavior, affecting both the UAV’s stability and the UGV’s traction due to the dynamic forces exerted by the tether. This behavior can be better appreciated in the accompanying video\footnote{\url{https://youtu.be/_j3KnPfFLBQ}}.

The ability to simulate these contact events is essential for realistic modeling of marsupial robotic systems, as tether-environment interactions can introduce significant constraints in real-world applications. Our results demonstrate that the simulator correctly captures these effects, enabling detailed analysis of how tether dynamics impact system performance under various operational conditions.

\subsection{Simulated Scenarios}

\begin{figure}[t!]
    \centering
     \includegraphics[width=1.0\columnwidth]{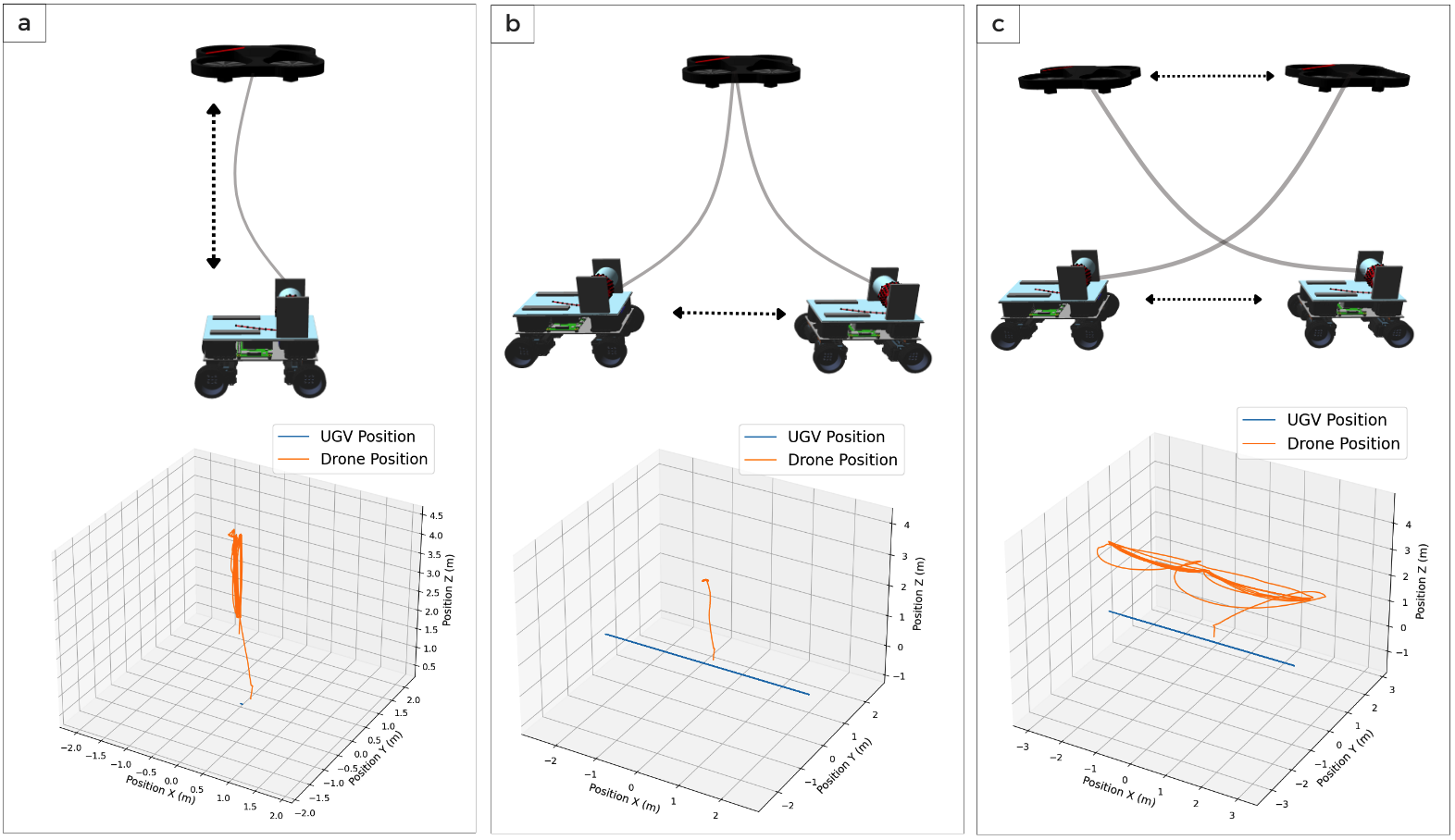}
    \caption{Examples of the scenarios used for the validation of the UAV-UGV tethered system simulator. (a) UAV ascending and descending with a static UGV. (b) UGV moving horizontally while the UAV remains stationary. (c) UAV and UGV moving in opposite directions.}
    \label{figure:tests_examples}
\end{figure}

In the simulated scenarios, we focus on evaluating the fundamental functioning of the simulation framework and analyzing the dynamics of the UAV and UGV when following predefined trajectories. These are designed to cover a range of fundamental operational conditions, allowing us to calibrate the simulator and identify any issues in basic tether management and vehicle control.

\begin{itemize}

\item \textbf{Scenario 1} (Fig. ~\ref{figure:tests_examples}a): \emph{Vertical Stability Assessment}. In this scenario, the UGV remains stationary while the UAV performs ten consecutive ascents and descents. The objective is to evaluate the winch's ability to smoothly release and retract the tether in response to vertical movements of the UAV. This scenario focuses on the stability of the UAV during tether length adjustments and the responsiveness of the winch control system.

\item \textbf{Scenario 2} (Fig.~\ref{figure:tests_examples}b): \emph{Horizontal Mobility Assessment}. The UAV hovers at a fixed altitude while the UGV moves back and forth between two points ten times. This scenario tests the tether's behavior during horizontal displacement of the UGV and examines the system's capability to manage tether slack without compromising the UAV's position stability.

\item \textbf{Scenario 3} (Fig.~\ref{figure:tests_examples}c): \emph{Opposite Direction Coordination}. In this scenario, both the UAV and UGV move simultaneously in opposite directions ten times. The aim is to challenge the system's coordination mechanisms and the winch's ability to adjust the tether length dynamically under increased complexity. This scenario simulates more intrincate movements that may occur in real-world operations where both units need to maneuver independently.

\end{itemize}

According to Table~\ref{tab:simulated_test_metrics}, the UGV traveled significant distances in Scenarios 2 and 3, covering 49.99 meters and 50.88 meters, respectively. The UGV consistently adhered to the designated paths, exhibiting smooth and stable movements that reflect reliable ground dynamics.

In contrast, the UAV experienced minor perturbations in its flight path, particularly during complex maneuvers or when operating in close proximity to the UGV. These disturbances are primarily attributed to the complex dynamics of the tether. The interactions among the tether's elements introduce dynamic forces that influence the UAV's stability and control. The most significant perturbations occur when the UAV and UGV are very close to each other or aligned vertically, as the reduced spatial separation amplifies the interaction forces from the tether. This effect is evident in Fig.~\ref{figure:tests_examples}c, where the UAV's trajectory shows deviations during such configurations.

\begin{figure}[t!]
    \centering
    \includegraphics[width=0.49\columnwidth]{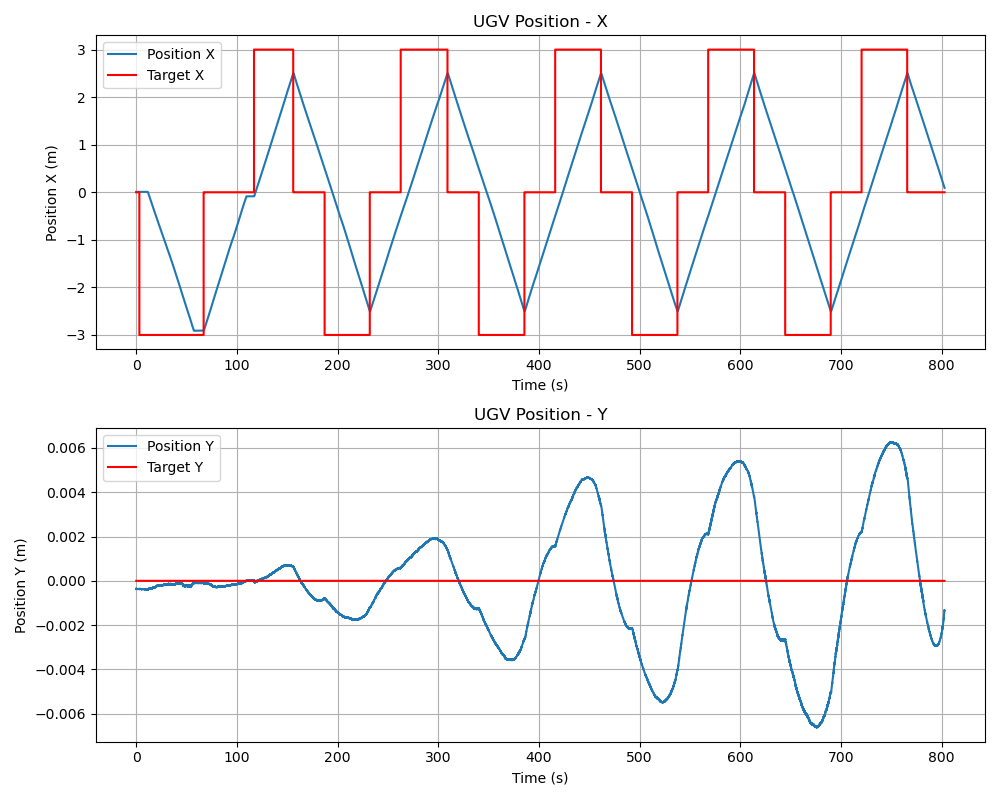}
    \includegraphics[width=0.49\columnwidth]{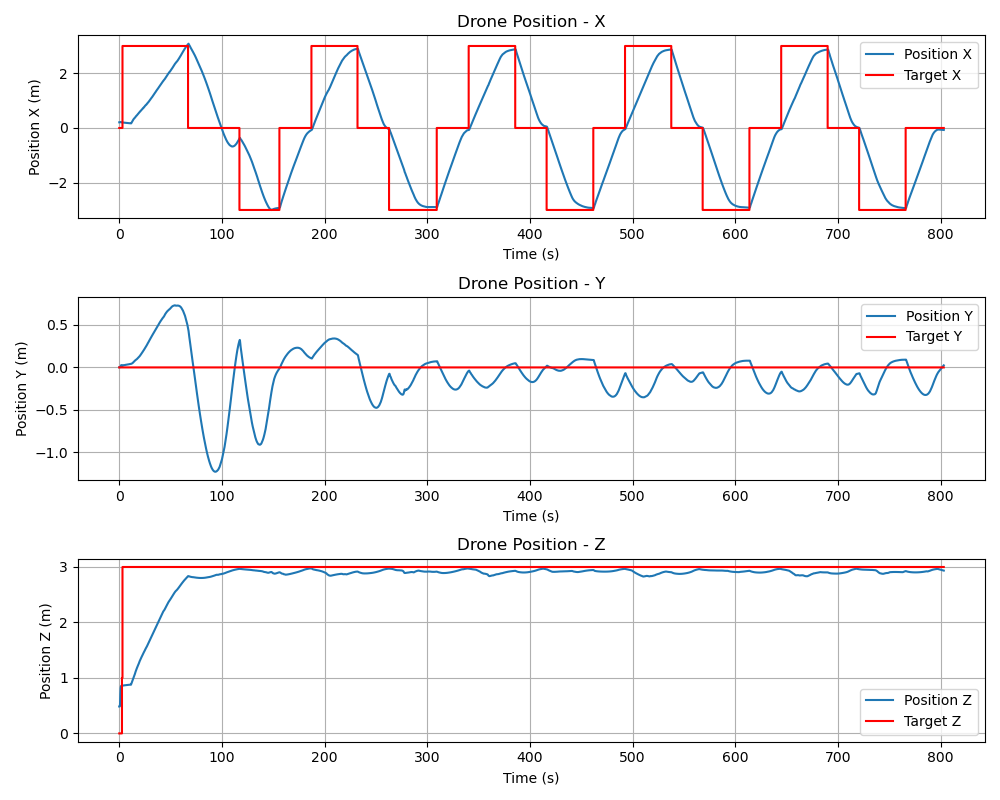}
    \caption{Visualization of the position of the UGV (left) and UAV (right) on each axis with respect to the reference in Scenario 3 (Opposite Direction Coordination).}
    \label{figure:test_position_ugv_uav}
\end{figure}

Notably, Scenario 3 resulted in the UAV traveling the greatest distance among the scenarios, covering 65.21 meters. The discrepancies between the UAV's actual path and the reference trajectory are further illustrated in Fig.~\ref{figure:test_position_ugv_uav}, which displays the UAV's and UGV's positions relative to their targets over time, highlighting slight oscillations due to the tether's influence.

Specifically, the table reports key performance metrics: the simulation time (in seconds), the number of targets reached, the total distance traveled by the UAV and UGV (in meters), and the amount of tether released and collected (in meters) throughout the experiment

\begin{table}[t!]
\centering
\caption{Simulation results of the simulated experiment scenario. Key performance metrics include simulation time (in seconds), number of targets reached, total distance traveled by the UAV and UGV (in meters), and the amount of tether released and collected (in meters) throughout the experiment.} \label{tab:simulated_test_metrics} 
\begin{tabular}{cccccccc}
\toprule
\textbf{Scenario} & \textbf{\makecell{Simulation \\ time}} & \textbf{\makecell{Number \\ of targets}} & \textbf{\makecell{Distance \\ UAV}} & \textbf{\makecell{Distance \\ UGV}} & \textbf{\makecell{Tether \\ released}} & \textbf{\makecell{Tether \\ collected}}   \\ 
\midrule
\textbf{1} & 504 & 20  & 45.50 & 0.25  & 23.64 & 22.08 \\        
\textbf{2} & 765 & 20  & 6.99  & 49.99 & 14.09 & 11.65 \\        
\textbf{3} & 803 & 20  & 65.21 & 50.88 & 40.11 & 37.68 \\      
\botrule
\end{tabular}
\end{table}

Despite these challenges, the UAV successfully reached all predefined target points across the stage, demonstrating the robustness of the control algorithms and the effectiveness of the tether management system. The control system effectively mitigated the disturbances induced by the tether, allowing the UAV to maintain its overall trajectory.

\begin{figure}[t!]
    \centering
    \includegraphics[width=0.6\columnwidth]{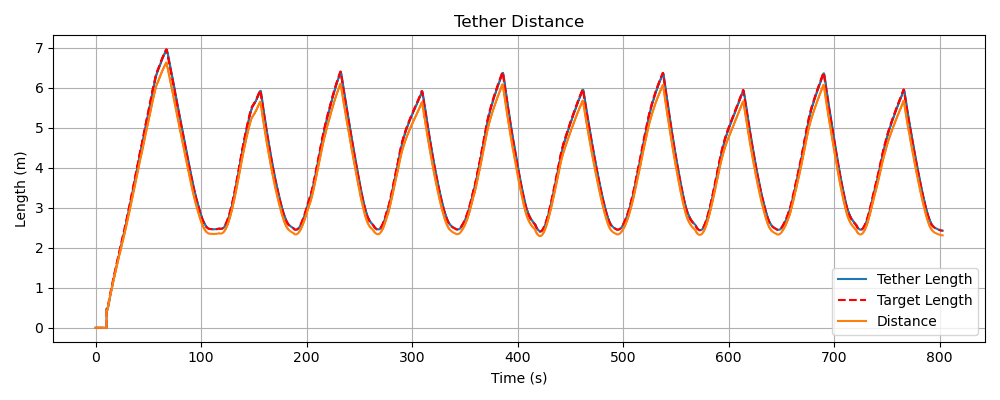}
    \includegraphics[width=0.6\columnwidth]{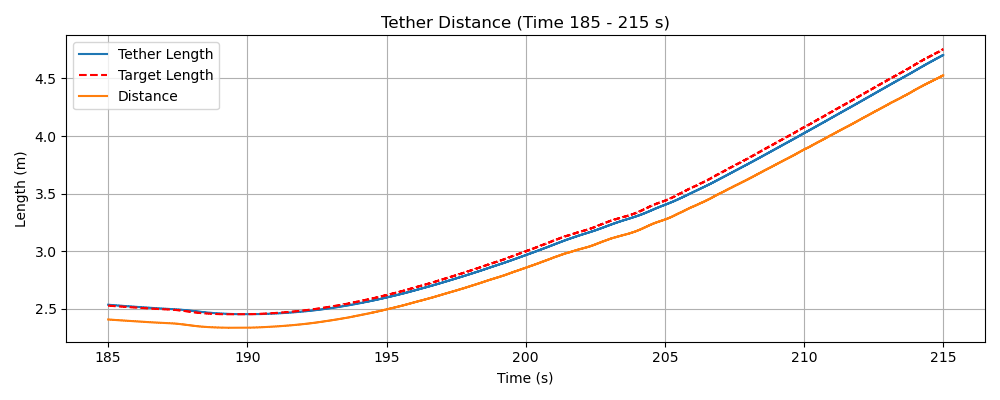}
    \caption{Tether length adjustments during Scenario 3, showing the evolution of the tether's actual length, target length, and distance between the UAV and UGV over time.}
    \label{figure:test_tether}
\end{figure}

Figure~\ref{figure:test_tether} depicts the tether length adjustments throughout the simulated scenario. The tether length is dynamically adjusted by the winch system to be 5\% greater than the relative distance between the UAV and UGV. This strategy ensures that the tether maintains slightly slack, preventing excessive tightness that might impact the UAV's stability. The tether length closely follows the target length, indicating the winch's responsiveness and the effectiveness of the tether management algorithms in real-time operation.

Overall, the simulated scenarios confirm that while the tether introduces additional complexity into the UAV's dynamics, the system is capable of maintaining accurate trajectory tracking.

\subsection{Comparison with real experiment}

\begin{figure}[t!]
    \centering
    \includegraphics[width=1.0\columnwidth]{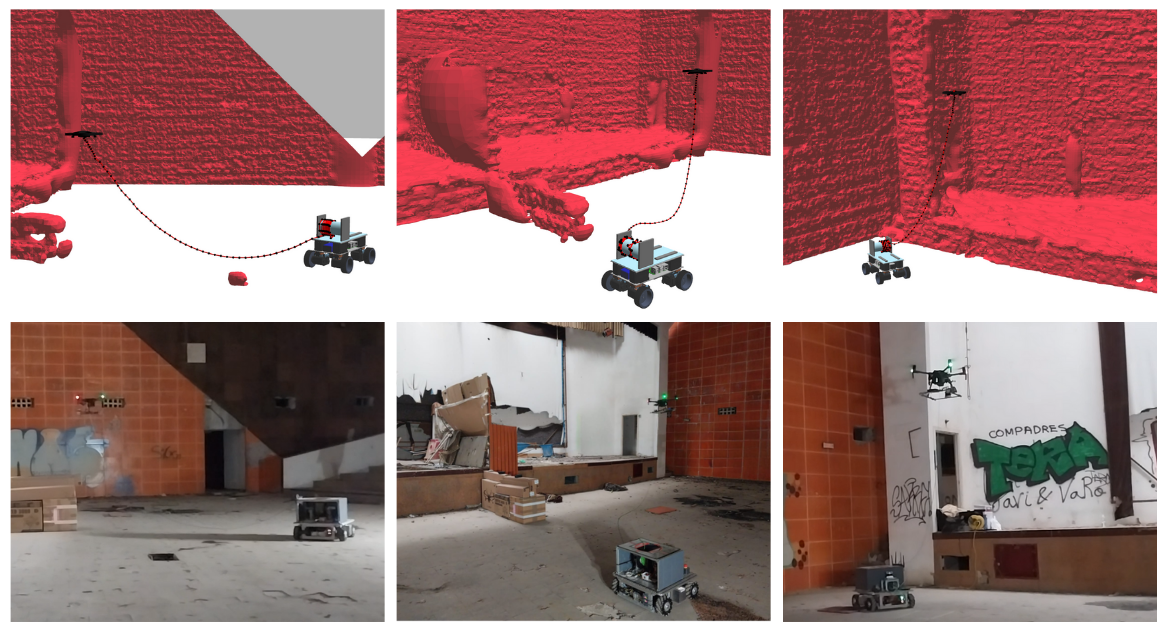}
    \caption{3D Lidar reconstruction of the theater in Gazebo where the experiment was conducted (up) and the real-world experiment used as a reference to validate the proposed simulation framework (down).}
    \label{figure:teatro_escenario}
\end{figure}

To evaluate the realism and accuracy of the proposed simulator, a real-world experiment with a marsupial robotic system was first conducted by our team (see Fig.~\ref{figure:teatro_escenario}), then replicated in simulation, and subsequently compared to the real experiment. This experiment, detailed in \cite{Martinez-Rozas2023}, was conducted with a physical tethered UAV-UGV system navigating a confined environment, a theatre. The marsupial system consisted of a UAV and a UGV connected by an adjustable hanging tether, with both the UAV and the UGV actively navigating, while the UGV also provided power and managed the tether. The experiment demonstrated the feasibility of planning and executing collision-free trajectories in a real-world setting, highlighting the challenges posed by tether dynamics and confined spaces.

Using the same parameters and trajectory plans as those employed in the real-world experiment, we replicated the setup in our simulator. The goal was to assess the simulator's ability to model the behavior of the tether and the coordinated movements of the UAV and UGV in a highly constrained environment. The theatre environment was recreated in Gazebo using a Lidar reconstruction, including obstacles and the same initial conditions, as shown in Fig.~\ref{figure:teatro_escenario}.

To further evaluate the simulator's performance, two specific scenarios were designed based on the replicated environment. In the first scenario, Real-Time Tether Adaptation (RTTA), the tether length was dynamically adjusted throughout the mission proportionally to the distance between the vehicles, adapting in real time to the relative positions of the UAV and UGV. This approach allowed for continuous adjustments to ensure proper slack and operational stability. In the second scenario, Predefined Tether Reference (PTR), the same trajectory was replicated using a pre-planned tether length for each waypoint of the trajectory, mirroring the behaviour of the real-world system, where the tether length is planned as part of the trajectory. 


The comparison shown in Table~\ref{tab:real_test_metrics} between these two scenarios offered valuable insights into how different tether management strategies affect system behavior and synchronization. The values for tether released and collected differ between the RTTA and PTR scenarios due to their distinct tether management strategies. Notably, the greatest discrepancies between simulation and real-world experiments occur when the UAV and UGV are further apart, as the extended tether introduces additional dynamic effects that influence the UAV’s motion. However, these differences remain minimal, and the overall trajectory alignment demonstrates the simulator’s ability to accurately capture key aspects of real-world marsupial robotic behavior. The UAV and UGV successfully navigate the complex environment, negotiating obstacles and coordinating movements effectively. This analysis underscored the versatility and fidelity of the simulator in modeling complex marsupial robotic systems.

\begin{figure}[]
    \centering
    \begin{subfigure}{0.48\columnwidth}
        \centering
        \includegraphics[width=\linewidth]{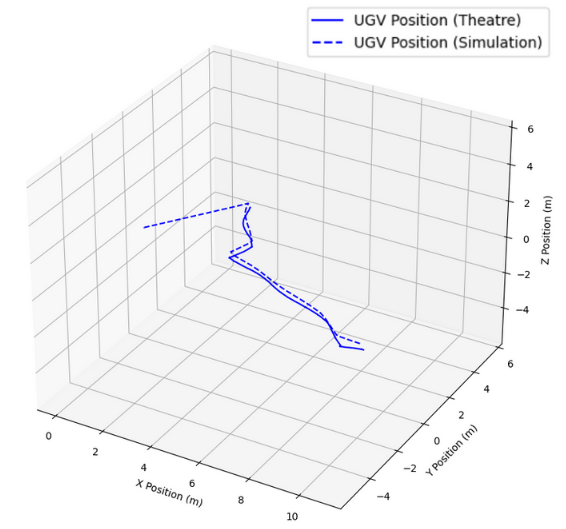}
        \caption{UGV comparison}
        \label{fig:ugv_comparison}
    \end{subfigure}
    \hfill
    \begin{subfigure}{0.48\columnwidth}
        \centering
        \includegraphics[width=\linewidth]{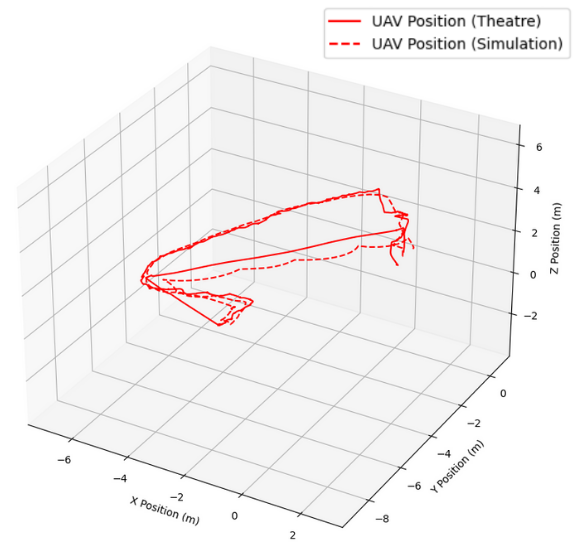}
        \caption{UAV comparison}
        \label{fig:uav_comparison}
    \end{subfigure}
    \caption{3D position comparison: UGV and UAV real theatre experiment vs simulation scenario.}
    \label{figure:test_theatre_3d}
\end{figure}

Figure~\ref{figure:test_theatre_3d} shows the 3D trajectories of the UAV and UGV during PTR scenario. Comparative analysis reveals that the deviations between the simulated and real-world trajectories are marginal. The UAV maintains stability despite the complex tether dynamics, and the UGV follows the planned path with high fidelity. The tether management system effectively adjusts the tether length in real-time, preventing excessive slack or tightness that could impede the robots' movements or compromise safety.

By analyzing the results from these scenarios, we can identify strengths and limitations of the simulation framework, ensuring its reliability for developing and testing algorithms for tethered marsupial robotic systems.

\begin{table}[t!]
\centering
\caption{Simulation results of the real experiment scenario. Key performance metrics include simulation time (in seconds), number of targets reached, total distance traveled by the UAV and UGV (in meters), and the amount of tether released and collected (in meters) throughout the experiment.} \label{tab:real_test_metrics} 
\begin{tabular}{cccccccc}
\toprule
\textbf{Scenario} & \textbf{\makecell{Simulation \\ time}} & \textbf{\makecell{Number \\ of targets}} & \textbf{\makecell{Distance \\ UAV}} & \textbf{\makecell{Distance \\ UGV}} & \textbf{\makecell{Tether \\ released}} & \textbf{\makecell{Tether \\ collected}}   \\ 
\midrule
\textbf{RTTA} & 628 & 100 & 33.92 & 16.45 & 17.91 & 7.54  \\       
\textbf{PTR} & 609 & 100 & 33.96 & 16.41 & 19.57 & 8.40  \\
\botrule
\end{tabular}
\end{table}

\subsection{Performance Evaluation}


The computational performance of the simulator was evaluated to assess its real-time capabilities and scalability. All experiments were conducted on a laptop with 32 GB of RAM, a 13th Gen Intel Core i7-13620H (10 cores, 16 threads), and an NVIDIA GeForce RTX 4060 Laptop GPU. To analyze the impact of system components on simulation efficiency, a comparison was made with a more powerful desktop PC (64 GB RAM, 12th Gen Intel Core i9-12900F, and an NVIDIA GeForce RTX 3060 GPU). Each experiment was repeated 10 times, and since the mean deviation was negligible, the values correspond to the calculated averages.

\begin{figure}[]
    \centering
    \includegraphics[width=0.9\columnwidth]{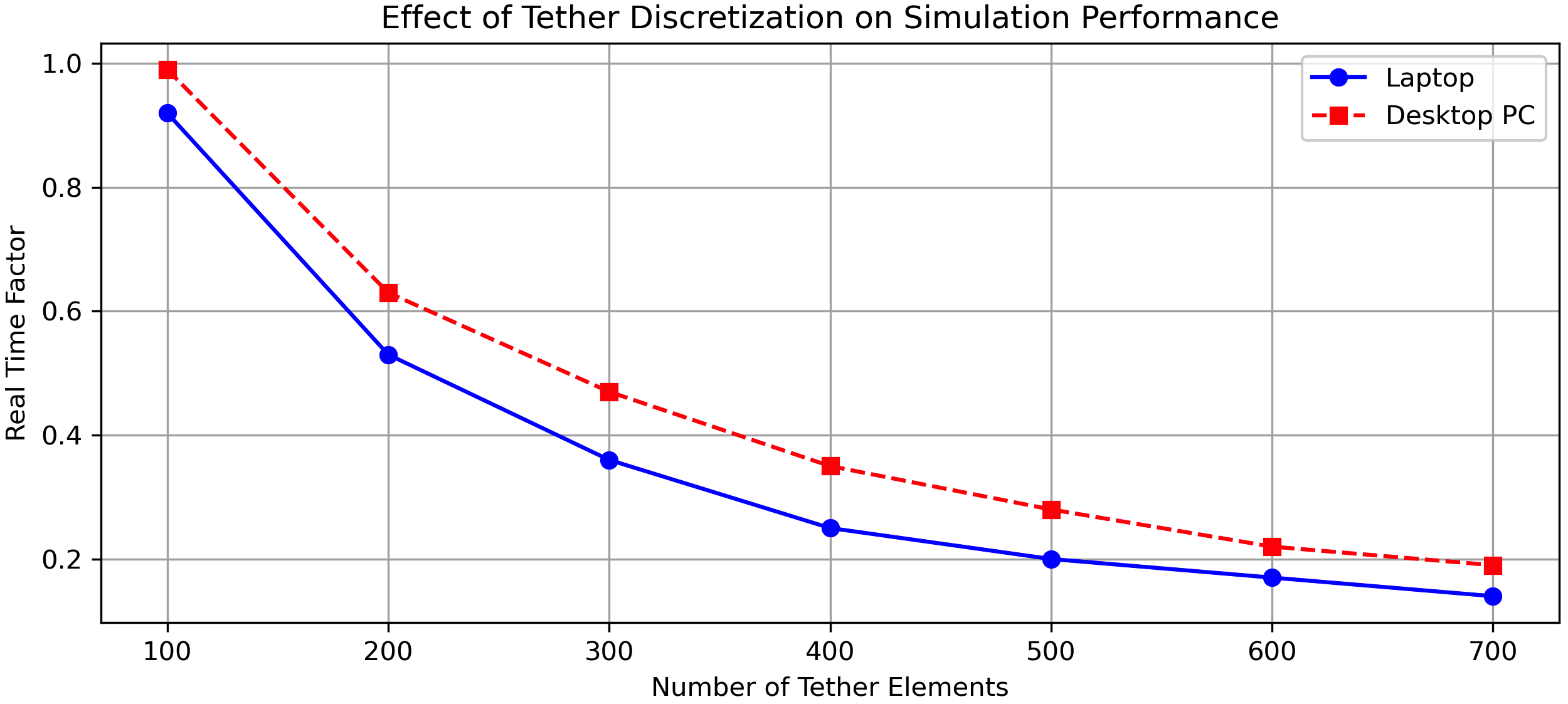}
    \caption{Performance comparison between the laptop and desktop configurations. The real-time factor (RTF) is shown for tether element counts ranging from 100 to 700.}
    \label{figure:performance_evaluation}
\end{figure}

As shown in Fig. \ref{figure:performance_evaluation}, the desktop PC demonstrated superior performance. The number of tether elements directly influenced computational load. For the laptop, the real-time factor (RTF) decreased from 0.92 (100 elements) to 0.14 (700 elements), while the desktop PC maintained higher RTF values, ranging from 0.99 (100 elements) to 0.19 (700 elements), with a tether element length of 0.15 m in all cases. This divergence underscores the importance of system specifications for simulating high-fidelity tethered systems. Notably, the desktop achieved an RTF of 0.47 with 300 elements while the laptop required element counts below 200 to maintain RTF \>0.50.

These results highlight the simulator's ability to run on mid-tier hardware while showcasing the benefits of higher-end systems for demanding simulations. As the number of tether elements increases, maintaining real-time performance becomes progressively more demanding. While the simulator remains functional across a range of systems, achieving high real-time factors, which is necessary if real-time execution is required, demands sufficient computational resources. Therefore, selecting the appropriate hardware is crucial to balancing simulation accuracy and computational feasibility in real-time applications.


\section{Conclusions and Future Work}
\label{sec:conclusions}

The experiments demonstrate the simulator's capability to accurately model the dynamics of tethered UAV-UGV systems. The UAV's minor trajectory disturbances, observed in the simulated scenarios, are consistent with expected behavior due to the complex interactions within the tether system. These disturbances are most pronounced when the UAV and UGV are in close proximity, highlighting the importance of careful tether management in such scenarios.

The successful replication of complex trajectories from real-world experiments highlights the simulator's effectiveness as a tool for research and development. The close alignment between the simulated and actual trajectories validates the physical models and control algorithms implemented within the simulator. This fidelity is crucial for researchers seeking to test and refine their algorithms in a risk-free environment before deploying them in real-world applications.

Moreover, the simulator's ability to handle both dynamic and predefined tether length adjustments provides flexibility in testing various tether management strategies. The consistent performance across different tether control approaches indicates that the simulator can accommodate a wide range of experimental setups and objectives.

Future work will focus on enhancing the winch model to make it more adaptable for integration with diverse ground vehicles, while improving the software architecture to support these modifications efficiently. Additionally, efforts will address complex tether dynamics, such as nonlinear effects and entanglement during high-speed maneuvers, to further improve the simulator’s fidelity and enable more accurate

\backmatter








\bibliography{sn-bibliography}


\begin{thebibliography}{11}
\ifx \bisbn   \undefined \def \bisbn  #1{ISBN #1}\fi
\ifx \binits  \undefined \def \binits#1{#1}\fi
\ifx \bauthor  \undefined \def \bauthor#1{#1}\fi
\ifx \batitle  \undefined \def \batitle#1{#1}\fi
\ifx \bjtitle  \undefined \def \bjtitle#1{#1}\fi
\ifx \bvolume  \undefined \def \bvolume#1{\textbf{#1}}\fi
\ifx \byear  \undefined \def \byear#1{#1}\fi
\ifx \bissue  \undefined \def \bissue#1{#1}\fi
\ifx \bfpage  \undefined \def \bfpage#1{#1}\fi
\ifx \blpage  \undefined \def \blpage #1{#1}\fi
\ifx \burl  \undefined \def \burl#1{\textsf{#1}}\fi
\ifx \doiurl  \undefined \def \doiurl#1{\url{https://doi.org/#1}}\fi
\ifx \betal  \undefined \def \betal{\textit{et al.}}\fi
\ifx \binstitute  \undefined \def \binstitute#1{#1}\fi
\ifx \binstitutionaled  \undefined \def \binstitutionaled#1{#1}\fi
\ifx \bctitle  \undefined \def \bctitle#1{#1}\fi
\ifx \beditor  \undefined \def \beditor#1{#1}\fi
\ifx \bpublisher  \undefined \def \bpublisher#1{#1}\fi
\ifx \bbtitle  \undefined \def \bbtitle#1{#1}\fi
\ifx \bedition  \undefined \def \bedition#1{#1}\fi
\ifx \bseriesno  \undefined \def \bseriesno#1{#1}\fi
\ifx \blocation  \undefined \def \blocation#1{#1}\fi
\ifx \bsertitle  \undefined \def \bsertitle#1{#1}\fi
\ifx \bsnm \undefined \def \bsnm#1{#1}\fi
\ifx \bsuffix \undefined \def \bsuffix#1{#1}\fi
\ifx \bparticle \undefined \def \bparticle#1{#1}\fi
\ifx \barticle \undefined \def \barticle#1{#1}\fi
\bibcommenthead
\ifx \bconfdate \undefined \def \bconfdate #1{#1}\fi
\ifx \botherref \undefined \def \botherref #1{#1}\fi
\ifx \url \undefined \def \url#1{\textsf{#1}}\fi
\ifx \bchapter \undefined \def \bchapter#1{#1}\fi
\ifx \bbook \undefined \def \bbook#1{#1}\fi
\ifx \bcomment \undefined \def \bcomment#1{#1}\fi
\ifx \oauthor \undefined \def \oauthor#1{#1}\fi
\ifx \citeauthoryear \undefined \def \citeauthoryear#1{#1}\fi
\ifx \endbibitem  \undefined \def \endbibitem {}\fi
\ifx \bconflocation  \undefined \def \bconflocation#1{#1}\fi
\ifx \arxivurl  \undefined \def \arxivurl#1{\textsf{#1}}\fi
\csname PreBibitemsHook\endcsname

\bibitem[\protect\citeauthoryear{Martínez-Rozas et~al.}{2023}]{Martinez-Rozas2023}
\begin{botherref}
\oauthor{\bsnm{Martínez-Rozas}, \binits{S.}},
\oauthor{\bsnm{Alejo}, \binits{D.}},
\oauthor{\bsnm{Caballero}, \binits{F.}},
\oauthor{\bsnm{Merino}, \binits{L.}}:
Path and trajectory planning of a tethered UAV-UGV marsupial robotic system
(2023).
\url{https://arxiv.org/abs/2204.01828}
\end{botherref}
\endbibitem

\bibitem[\protect\citeauthoryear{Papachristos and Tzes}{2014}]{Papachristos2014}
\begin{bchapter}
\bauthor{\bsnm{Papachristos}, \binits{C.}},
\bauthor{\bsnm{Tzes}, \binits{A.}}:
\bctitle{The power-tethered uav-ugv team: A collaborative strategy for navigation in partially-mapped environments}.
In: \bbtitle{22nd Mediterranean Conference on Control and Automation},
pp. \bfpage{1153}--\blpage{1158}
(\byear{2014}).
\doiurl{10.1109/MED.2014.6961531}
\end{bchapter}
\endbibitem

\bibitem[\protect\citeauthoryear{Hudson et~al.}{2022}]{Hudson2022}
\begin{barticle}
\bauthor{\bsnm{Hudson}, \binits{N.}},
\bauthor{\bsnm{Talbot}, \binits{F.}},
\bauthor{\bsnm{Cox}, \binits{M.}},
\bauthor{\bsnm{Williams}, \binits{J.}},
\bauthor{\bsnm{Hines}, \binits{T.}},
\bauthor{\bsnm{Pitt}, \binits{A.}},
\bauthor{\bsnm{Wood}, \binits{B.}},
\bauthor{\bsnm{Frousheger}, \binits{D.}},
\bauthor{\bsnm{Lo~Surdo}, \binits{K.}},
\bauthor{\bsnm{Molnar}, \binits{T.}},
\bauthor{\bsnm{Steindl}, \binits{R.}},
\bauthor{\bsnm{Wildie}, \binits{M.}},
\bauthor{\bsnm{Sa}, \binits{I.}},
\bauthor{\bsnm{Kottege}, \binits{N.}},
\bauthor{\bsnm{Stepanas}, \binits{K.}},
\bauthor{\bsnm{Hernandez}, \binits{E.}},
\bauthor{\bsnm{Catt}, \binits{G.}},
\bauthor{\bsnm{Docherty}, \binits{W.}},
\bauthor{\bsnm{Tidd}, \binits{B.}},
\bauthor{\bsnm{Tam}, \binits{B.}},
\bauthor{\bsnm{Murrell}, \binits{S.}},
\bauthor{\bsnm{Bessell}, \binits{M.}},
\bauthor{\bsnm{Hanson}, \binits{L.}},
\bauthor{\bsnm{Tychsen-Smith}, \binits{L.}},
\bauthor{\bsnm{Suzuki}, \binits{H.}},
\bauthor{\bsnm{Overs}, \binits{L.}},
\bauthor{\bsnm{Kendoul}, \binits{F.}},
\bauthor{\bsnm{Wagner}, \binits{G.}},
\bauthor{\bsnm{Palmer}, \binits{D.}},
\bauthor{\bsnm{Milani}, \binits{P.}},
\bauthor{\bsnm{O’Brien}, \binits{M.}},
\bauthor{\bsnm{Jiang}, \binits{S.}},
\bauthor{\bsnm{Chen}, \binits{S.}},
\bauthor{\bsnm{Arkin}, \binits{R.}}:
\batitle{Heterogeneous ground and air platforms, homogeneous sensing: Team csiro data61’s approach to the darpa subterranean challenge}.
\bjtitle{Field Robotics}
\bvolume{2}(\bissue{1}),
\bfpage{595}--\blpage{636}
(\byear{2022})
\doiurl{10.55417/fr.2022021}
\end{barticle}
\endbibitem

\bibitem[\protect\citeauthoryear{Oh et~al.}{2006}]{Oh2006}
\begin{barticle}
\bauthor{\bsnm{Oh}, \binits{S.-R.}},
\bauthor{\bsnm{Pathak}, \binits{K.}},
\bauthor{\bsnm{Agrawal}, \binits{S.K.}},
\bauthor{\bsnm{Pota}, \binits{H.R.}},
\bauthor{\bsnm{Garratt}, \binits{M.}}:
\batitle{Approaches for a tether-guided landing of an autonomous helicopter}.
\bjtitle{IEEE Transactions on Robotics}
\bvolume{22}(\bissue{3}),
\bfpage{536}--\blpage{544}
(\byear{2006})
\doiurl{10.1109/TRO.2006.870657}
\end{barticle}
\endbibitem

\bibitem[\protect\citeauthoryear{Bulić et~al.}{2022}]{Bulic2022}
\begin{barticle}
\bauthor{\bsnm{Bulić}, \binits{D.}},
\bauthor{\bsnm{Tolić}, \binits{D.}},
\bauthor{\bsnm{Palunko}, \binits{I.}}:
\batitle{Beam-based tether dynamics and simulations using finite element model}.
\bjtitle{IFAC-PapersOnLine}
\bvolume{55}(\bissue{15}),
\bfpage{154}--\blpage{159}
(\byear{2022})
\doiurl{10.1016/j.ifacol.2022.07.624} .
\bcomment{6th IFAC Conference on Intelligent Control and Automation Sciences ICONS 2022}
\end{barticle}
\endbibitem

\bibitem[\protect\citeauthoryear{Huang}{2018}]{Huang2023}
\begin{botherref}
\oauthor{\bsnm{Huang}, \binits{Y.}}:
sitl\_gazebo.
GitHub
(2018).
\url{https://github.com/RigidWing/sitl_gazebo}
\end{botherref}
\endbibitem

\bibitem[\protect\citeauthoryear{Caruso et~al.}{2021}]{Caruso2021}
\begin{botherref}
\oauthor{\bsnm{Caruso}, \binits{M.}},
\oauthor{\bsnm{Gallina}, \binits{P.}},
\oauthor{\bsnm{Seriani}, \binits{S.}}:
On the modelling of tethered mobile robots as redundant manipulators.
Robotics
\textbf{10}(2)
(2021)
\doiurl{10.3390/robotics10020081}
\end{botherref}
\endbibitem

\bibitem[\protect\citeauthoryear{Laranjeira et~al.}{2020}]{Laranjeira2020}
\begin{barticle}
\bauthor{\bsnm{Laranjeira}, \binits{M.}},
\bauthor{\bsnm{Dune}, \binits{C.}},
\bauthor{\bsnm{Hugel}, \binits{V.}}:
\batitle{Catenary-based visual servoing for tether shape control between underwater vehicles}.
\bjtitle{Ocean Engineering}
\bvolume{200},
\bfpage{107018}
(\byear{2020})
\doiurl{10.1016/j.oceaneng.2020.107018}
\end{barticle}
\endbibitem

\bibitem[\protect\citeauthoryear{Petit and Desbiens}{2022}]{Petit2022}
\begin{barticle}
\bauthor{\bsnm{Petit}, \binits{L.}},
\bauthor{\bsnm{Desbiens}, \binits{A.L.}}:
\batitle{Tape: Tether-aware path planning for autonomous exploration of unknown 3d cavities using a tangle-compatible tethered aerial robot}.
\bjtitle{IEEE Robotics and Automation Letters}
\bvolume{7}(\bissue{4}),
\bfpage{10550}--\blpage{10557}
(\byear{2022})
\doiurl{10.1109/LRA.2022.3194691}
\end{barticle}
\endbibitem

\bibitem[\protect\citeauthoryear{Luqman}{2024}]{Luqman2024}
\begin{botherref}
\oauthor{\bsnm{Luqman}, \binits{M.}}:
sjtu\_drone.
GitHub
(2024).
\url{https://github.com/noshluk2/sjtu_drone/tree/ros2}
\end{botherref}
\endbibitem

\bibitem[\protect\citeauthoryear{robot\_mania}{2023}]{4WSrobot}
\begin{botherref}
\oauthor{\bsnm{robot\_mania}}:
Simulation of a 4WS Robot Using ROS2 Control and Gazebo
(2023).
\url{https://www.youtube.com/watch?v=VX53gAXafUA}
\end{botherref}
\endbibitem

\end{thebibliography}
\end{document}